\documentclass[journal]{./sty/IEEEtran}
\usepackage{graphicx}
\usepackage{cite}
\usepackage{amsmath}
\usepackage{url}
\usepackage{amssymb}
\usepackage{tabularx}
\usepackage{multirow}
\usepackage{threeparttable}
\usepackage{caption}
\usepackage{booktabs}
\usepackage{ulem}
\usepackage{bbding}
\usepackage[ruled,linesnumbered]{algorithm2e}
\usepackage{color}
\usepackage[hidelinks]{hyperref}
\usepackage{amsthm,amsmath,amssymb,lipsum}
\usepackage{mathrsfs}
\usepackage{hyperref}
\usepackage{makecell}
\usepackage{tablefootnote}
\usepackage{float}
\usepackage{threeparttable}
\usepackage{afterpage}
\usepackage{ragged2e}
\usepackage{longtable}

\renewcommand{\arraystretch}{1.4}

\newcolumntype{L}[1]{>{\raggedright\arraybackslash}p{#1}}
\newcolumntype{C}[1]{>{\centering\arraybackslash}p{#1}}
\newcolumntype{R}[1]{>{\raggedleft\arraybackslash}p{#1}}

\newcommand{\etc}{\textit{etc}}
\newcommand{\etal}{\textit{et al.}}

\newcommand{\settablefont}{\fontsize{7.5}{8.6}\selectfont}

\begin{document}

\title{Vehicular Road Crack Detection with Deep Learning:  A New Online Benchmark for Comprehensive Evaluation of Existing Algorithms}

\author{Nachuan Ma, Zhengfei Song, Qiang Hu, Chuang-Wei Liu, Yu Han, Yanting Zhang,~\IEEEmembership{Member,~IEEE},
\\	Rui Fan,~\IEEEmembership{Senior Member,~IEEE}, and 
	Lihua Xie,~\IEEEmembership{Fellow,~IEEE }

\thanks{This research was supported in part by the National Key R\&D Program of China under Grant 2023YFE0202400, the Science and Technology Commission of Shanghai Municipal under Grant 22511104500, the Fundamental Research Funds for the Central Universities, and National Natural Science Foundation of China under Grant 62206046. \emph{(Corresponding author: Rui Fan)}}
\thanks{N. Ma, Z. Song, Q. Hu, C. Liu and R. Fan are with the College of Electronics \& Information Engineering, Shanghai Research Institute for Intelligent Autonomous Systems, the State Key Laboratory of Intelligent Autonomous Systems, and Frontiers Science Center for Intelligent Autonomous Systems, Tongji University, Shanghai 201804, China. (e-mail: 2111481@tongji.edu.cn, rui.fan@ieee.org) }
\thanks{Y. Han and Y. Zhang are with the School of Computer Science and Technology, Donghua University, Shanghai 201620, P. R. China. (e-mails: {2232816@mail.dhu.edu.cn, ytzhang@dhu.edu.cn}}
\thanks{L. Xie is with the School of Electrical and Electronic Engineering, Nanyang Technological University, 50 Nanyang Avenue, Singapore 639798 (e-mail: elhxie@ntu.edu.sg).}
}

\markboth{IEEE TRANSACTIONS ON INTELLIGENT Vehicles}{}

\maketitle
\begin{abstract}

In the emerging field of urban digital twins (UDTs), advancing intelligent road inspection (IRI) vehicles with automatic road crack detection systems is essential for maintaining civil infrastructure. Over the past decade, deep learning-based road crack detection methods have been developed to detect cracks more efficiently, accurately, and objectively, with the goal of replacing manual visual inspection. Nonetheless, there is a lack of systematic reviews on state-of-the-art (SoTA) deep learning techniques, especially data-fusion and label-efficient algorithms for this task. This paper thoroughly reviews the SoTA deep learning-based algorithms, including (1) supervised, (2) unsupervised, (3) semi-supervised, and (4) weakly-supervised methods developed for road crack detection. Also, we create a dataset called UDTIRI-Crack, comprising $2,500$ high-quality images from seven public annotated sources, as the first extensive online benchmark in this field. Comprehensive experiments are conducted to compare the detection performance, computational efficiency, and generalizability of public SoTA deep learning-based algorithms for road crack detection. In addition, the feasibility of foundation models and large language models (LLMs) for road crack detection is explored. Afterwards, the existing challenges and future development trends of deep learning-based road crack detection algorithms are discussed. We believe this review can serve as practical guidance for developing intelligent road detection vehicles with the next-generation road condition assessment systems. The released benchmark UDTIRI-Crack is available at https://udtiri.com/submission/.
\end{abstract}


\IEEEpeerreviewmaketitle

\begin{figure*}[!t]
	\begin{center}
		\centering
		\includegraphics[width=1\textwidth]{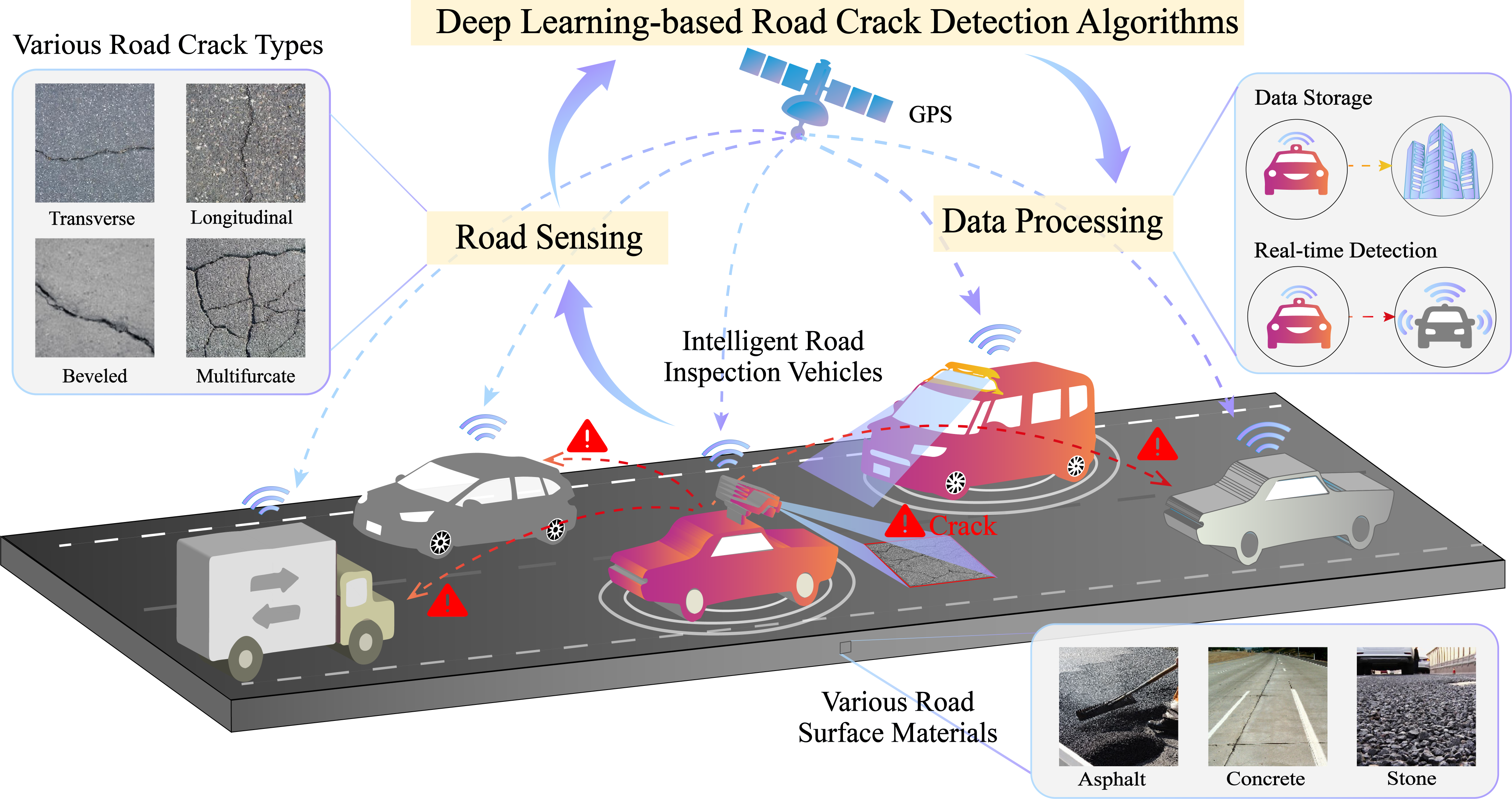}
		\centering
        \captionsetup{font={small}}
		\caption{Intelligent road inspection vehicles with automatic road crack detection systems.}
		\label{fig_intro}
	\end{center}
\end{figure*}

\section{Introduction}

\IEEEPARstart{C}{racks} are narrow, dark lines or curves that manifest on the surfaces of solid materials, such as roads and bridges \cite{zou2018deepcrack}. Road cracks emerge due to the combined effects of water and traffic-related factors \cite{miller2003distress}, such as soil swelling, foundation shifting, traffic congestion, and the expansion and contraction of materials. Beyond being inconveniences, road cracks substantially undermine the reliability and sustainability of civil infrastructure and pose significant threats to vehicle integrity and driving safety \cite{liu2019deepcrack}. For instance, in the United Kingdom, poor road surfaces accounted for $12.6\%$ of all traffic accidents in $2020$, according to the Department for Transport \cite{roadsafe}. Therefore, to lower the risk of structural degradation and traffic accidents, frequent road inspection is necessary \cite{fan2019crack}. Also, as autonomous vehicle driving becomes increasingly prevalent, it is crucial to ensure road surface health for safely operating of autonomous deployment.

Currently, manual visual inspection is still the dominant method for road crack detection \cite{fan2019road}. The locations of road cracks are recorded routinely by civil engineers or qualified inspectors, the process of which is time-consuming, costly, and hazardous \cite{fan2019pothole}. In countries like China or the United States, over $100,000$ kilometers of highways require regular testing and maintenance, which demands significant labor costs. Moreover, the detection results are always qualitative and subjective, as decisions depend entirely on personal opinions and expertise. Owing to these concerns, developing intelligent road inspection vehicles equipped with automatic road condition monitoring algorithms becomes an ever-increasing need \cite{fan2021rethinking}. As illustrated in Fig. \ref{fig_intro}, such vehicles have the potential to process road data and detect road cracks with high accuracy, efficiency, and objectivity, helping to reduce labor costs and further improve road maintenance efficiency through online analysis. Therefore, a new review that reflects the recent research trend of automated road crack detection algorithms is needed. 

Traditional computer vision-based road crack detection methods utilize image processing-based techniques, including edge-based \cite{zhao2015anisotropic, ayenu2008evaluating}, thresholding-based \cite{yamaguchi2008image}, texture analysis-based \cite{hu2010novel}, wavelet-based \cite{zhou2006wavelet}, and minimal path search-based \cite{amhaz2016automatic}. While these methods may demonstrate effectiveness in certain simple scenarios, they are sensitive to environmental factors like illumination and weather. The geometric assumptions used in such methods can also be impractical due to the irregular shapes of road cracks. 

Fortunately, recent advances in deep learning have led to the extensive use of convolutional neural networks (CNNs) and Transformer-based models for automated road crack detection. 
Unlike traditional methods with explicit parameters and hand-crafted features, these models use annotated road data to learn implicit parameters through back-propagation. They are categorized into three types: (1) image classification networks, which differentiate between crack and non-crack images \cite{krizhevsky2017imagenet, fan2021deep}; (2) object detection networks, which identify cracks at the instance level (location and class) \cite{cha2018autonomous, du2021pavement}; and (3) semantic segmentation networks, which perform pixel-wise crack detection and have become the preferred approach \cite{dung2019autonomous, huyan2020cracku, qu2021crack, chen2020pavement, liu2020automated, zhang2022intelligent, zhu2023lightweight}. Though semantic segmentation algorithms have shown great potential in accurately detecting cracks, training such algorithms requires extensive human-annotated datasets with meticulous annotations, demanding significant labor and time costs. Therefore, label-efficient deep learning-based methods have been developed, aiming to mitigate or even get rid of the dependence on fine annotated datasets, which can be divided into unsupervised \cite{ma2024up}, semi-supervised \cite{liu2024semi}, and weakly-supervised \cite{zhang2022investigation} methods. Also, in recent years, there has been increasing interest in using depth maps \cite{li2024cnn} \cite{jing2024self} for road crack detection, which can provide essential 3D information that complements image data, enhancing detection performance through data fusion. However, acquiring such data necessitates costly specialized equipment, and effectively integrating multi-modal information remains a challenge in algorithm development. 

This paper aims to thoroughly explore the existing deep learning-based algorithms for road crack detection and provide future trends for upcoming research. Besides reviewing algorithms, we curated a high-quality dataset called UDTIRI-Crack as an extended dataset of UDTIRI \cite{guo2024udtiri}, containing $2,500$ high-quality images selected from seven public datasets with various crack types and road surface materials, under diverse scenes and lighting conditions, which has been promoted as the first extensive online benchmark in this field. We conduct experiments using public general-purpose and crack detection-specific supervised methods on the proposed UDTIRI-Crack, AigleRN \cite{amhaz2016automatic}, and CrackNJ156 \cite{xu2022pavement} datasets to compare the detection accuracy, generalizability, model parameters, and processing speed of them. To the best of our knowledge, this study is the first to comprehensively compare the performance of advanced semantic segmentation-based algorithms in this field. In addition, we explore the feasibility of foundation models and large language models (LLMs) for road crack detection and conducted experiments using the Segment Anything Model (SAM) \cite{kirillov2023segment} series and Grounded-SAM \cite{ren2024grounded}. The results indicate that further efforts are needed to adapt them specifically for road crack detection.

\begin{figure*}[!t]
	\begin{center}
		\centering
		\includegraphics[width=1\textwidth]{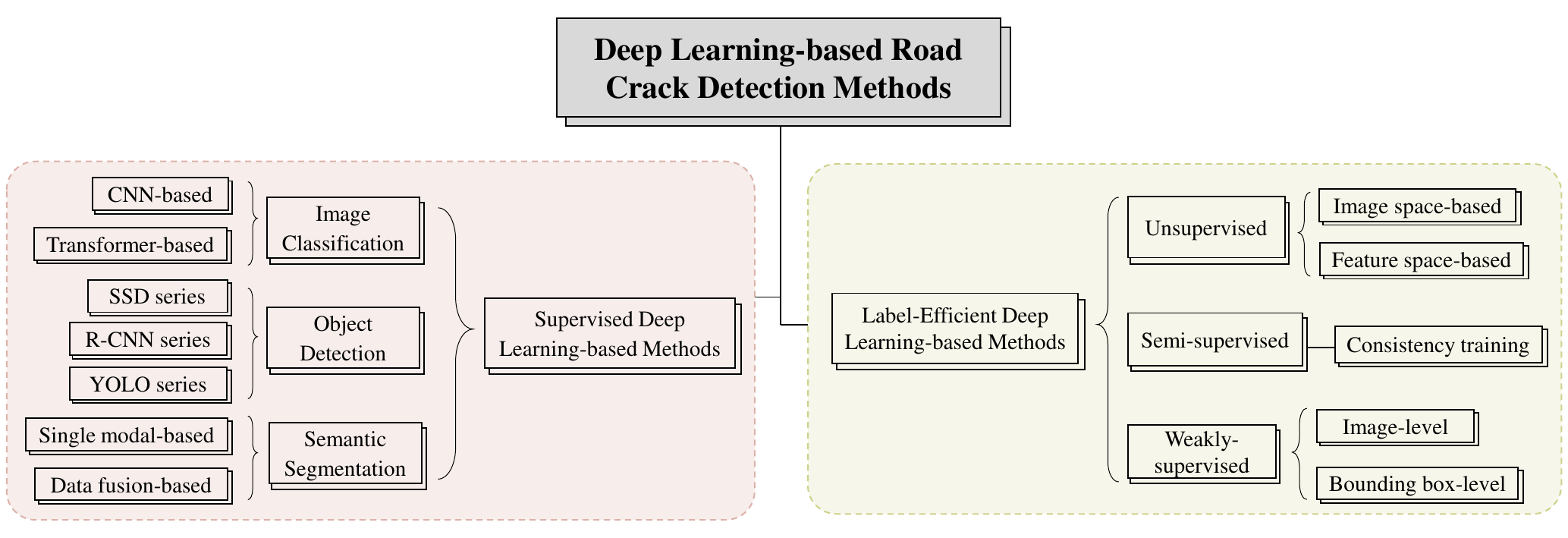}
        \captionsetup{font={small}}
		\caption{The overall outline of the reviewed computer deep learning-based road crack detection methods.}
		\label{fig_outline}
	\end{center}
\end{figure*}

The structure of this paper is in the following manner: Sect. \ref{sec.deep} centers on the advanced supervised deep learning-based road crack detection methods. Sect. \ref{sec.under} reviews the existing label-efficient deep learning-based methods. Sect. \ref{sec.experiment} summarizes public road crack datasets and presents comparison results of public semantic segmentation-based algorithms. Sect. \ref{sec.future} presents existing challenges and future trends. Sect. \ref{sec.conclusion} concludes the paper. The overall outline of the reviewed algorithms is shown in Fig. \ref{fig_outline}.

\begin{table*}[!t]
\begin{threeparttable}
\caption{List of selected image classification-based and object detection-based road crack detection methods}
\label{table_class2object}  
\begin{tabular}{R{1cm}R{2.4cm}R{0.6cm}C{2.8cm}m{9.1cm}}
\toprule
Type & Reference & Year & Key method & Description\\
\midrule
\multirow{4}{*}[-3em]{\makecell[c]{Image \\ Classification}} 
 & Kim \etal \cite{kim2018automated} & 2018 & AlexNet & Kim \etal \cite{kim2018automated} trained an enhanced version of AlexNet to classify road image regions, achieving good road crack detection performance. \\
 & Fan \etal \cite{fan2019crack} & 2019 & FCN & Fan \etal \cite{fan2019crack} introduced an FCN model for classifying crack images, supplemented by bilateral filtering and adaptive thresholding for crack area extraction. \\
 & Hou \etal \cite{hou2021mobilecrack} & 2021 & MobileNet & Hou \etal \cite{hou2021mobilecrack} developed a lightweight network for road crack detection, called MobileCrack, which surpassed MobileNet in performance while requiring roughly one-quarter of the computational load. \\
 & Chen \etal \cite{chen2022fast} & 2022 & LeViT & Chen \etal \cite{chen2022fast} trained a LeViT model \cite{graham2021levit} for rapid inference in road image classification, outperforming both Vision Transformer (ViT) \cite{dosovitskiy2020image} and ResNet \cite{resnet} in terms of detection performance and inference speed. \\
\midrule
\multirow{6}{*}[-6em]{\makecell[c]{Object \\ Detection}} 
 & Maeda \etal \cite{maeda2018road} & 2018 & SSD & Maeda \etal \cite{maeda2018road} incorporated Inception-v2 \cite{szegedy2016rethinking} and MobileNet \cite{howard2017mobilenets} as the backbone into SSD for road crack detection on road images. \\
 & Song \etal \cite{song2021faster} & 2021 & Faster R-CNN & Song \etal \cite{song2021faster} trained 20 different Faster R-CNN models with various anchor settings to determine the most suitable configuration for analyzing road images from highways in Xinjiang, China. \\
 & Pham \etal \cite{pham2020road} & 2022 & Faster R-CNN & An extensive comparison of Faster R-CNN models with different backbones and setups conducted by \cite{pham2020road} highlighted that the Faster R-CNN (with ResNetXt101 as the backbone) achieved good overall performance. \\
 & Ma \etal \cite{ma2022automatic} & 2022 & YOLOv3 & In Ma \etal \cite{ma2022automatic}, a Pavement Crack Generative Adversarial Network (PCGAN) was designed to synthesize realistic crack images for training an improved YOLOv3 network capable of rapid road crack detection in video frames. The proposed model facilitates its implementation in mobile devices and systems mounted on vehicles and UAVs, enabling real-time road crack detection. \\
 & Du \etal \cite{du2022improvement} & 2022 & YOLOv5 & Du \etal \cite{du2022improvement} proposed BV-YOLOv5 by integrating the bidirectional feature pyramid network and Varifocal loss into the YOLOv5, which boosted the network's detection capabilities and inference speed for road crack detection. \\
 & Su \etal \cite{su2024mod} & 2024 & MOD-YOLO & Su \etal \cite{su2024mod} introduced the Maintaining the Original Dimension-YOLO (MOD-YOLO) to address channel information loss and limited receptive fields in previous YOLO series for road crack detection. It incorporated MODSConv blocks to enhance inter-channel communication in the feature layers, as well as a global receptive field-space pooling pyramid module for improved network's global perception and scale adaptability. Comparisons show that MOD-YOLO outperforms YOLOX \cite{ge2021yolox} in accuracy, efficiency, and generalizability, with fewer parameters and lower computational complexity. \\
\bottomrule
\end{tabular} 
\end{threeparttable}
\end{table*}

\section{Supervised Deep Learning-based Road Crack Detection Methods}
\label{sec.deep}

Recent advances in deep learning have elevated deep convolutional neural networks (DCNNs) and Transformer networks as primary tools for road crack detection. Unlike traditional approaches that require explicit parameter settings, DCNNs and Transformer networks typically undergo training via back-propagation using extensive datasets of human-annotated road images \cite{lecun2015deep}. These data-driven road crack detection approaches encapsulate three principal modalities, as delineated in \cite{fan2023autonomous}: (1) image classification networks, (2) object detection networks, and (3) semantic segmentation networks. Here, image classification networks are devised to discern between images containing cracks (positive) and those without (negative), object detection networks aim to recognize cracks at the instance level, and semantic segmentation networks are trained to execute pixel-level (or semantic-level) detection of road images. The ensuing discussion elaborates on each of these algorithmic approaches, and the selected methods are summarized in Table \ref{table_class2object}, \ref{table_singlesemantic}, and Table \ref{table_datafusionsemantic}.

\subsection{Image Classification Networks}

Prior to the advent of deep learning technologies, researchers primarily employed classical image processing algorithms to generate hand-crafted visual features, which were then utilized to train machine learning models, such as SVM \cite{kapela2015asphalt, sensors, SVMOTSU, LBPPCASVM}, RF \cite{decision} \cite{shi2016automatic}, and AdaBoost (AB) \cite{cord2012automatic} to classify road image patches. Although these machine learning-based road crack detection methods may be effective under some simplistic conditions, they are limited by the inadequacy of feature selection and the labor-intensiveness of manual feature extraction efforts. Moreover, these methods are highly susceptible to noise and generally perform poorly when analyzing road images characterized by poor lighting conditions or complex road textures. 

The exceptional feature extraction ability of deep learning networks has promoted their widespread adoption in road crack classification methods \cite{cha2017deep,zhang2016road,fan2021deep,kim2018automated,fan2019crack,hou2021mobilecrack,chen2022fast}, including fully connected network (FCN) \cite{long2015fully}, Alexnet \cite{krizhevsky2017imagenet}, Xception \cite{chollet2017xception}, SENet \cite{SENet}, PNASNet \cite{liu2018progressive}, MobileNet \cite{howard2017mobilenets}, \etc. For instance, \cite{cha2017deep} \cite{zhang2016road} used a fixed-size sliding window to separate the entire image into blocks and obtained classification results by applying CNN-based models, showing better robustness and adaptability than using traditional edge-based methods and machine learning methods. An extensive comparison among 30 SoTA image classification CNNs by \cite{fan2021deep} highlighted that PNASNet provided a good balance between inference speed and detection accuracy. The aforementioned methods suggest the relative simplicity of road crack detection as compared to image classification tasks in other application domains. To obtain details of crack location information, the focus should be evolved to object detection-based and semantic segmentation-based algorithms.   

\subsection{Object Detection Networks}

Object detection-based road crack detection methods are designed to localize crack areas with bounding boxes, which can be grouped into three types: (1) single shot multi-box detector (SSD)-based, (2) region-based CNN (R-CNN) series-based, and (3) you only look once (YOLO) series-based. 

An SSD is constructed from two main elements: the backbone model and the SSD head. The former is typically a deep image classification network tasked with extracting visual features, whereas the latter contains additional convolutional layers for the generation of bounding boxes associated with object classes \cite{maeda2018road}. However, SSD utilizes predefined aspect ratios and scales for anchor boxes, which greatly restricts its detection performance and generalizability. 

Compared to SSD, R-CNN \cite{cha2018autonomous,song2021faster,pham2020road,wang2018road,maeda2018road,kortmann2020detecting,gou2019pavement,shen2020pavement} and YOLO series are more widely used for road crack detection. For instance, \cite{cha2018autonomous} deployed the faster R-CNN devised by \cite{ren2015faster} to detect multi-scale cracks in concrete images. \cite{wang2018road} compared the performance of two Faster R-CNNs (with ResNet-101 and ResNet-152 as the backbones, separately) for road crack detection on the dataset introduced in \cite{maeda2018road}. The experimental results indicated superior performance by the latter, likely due to its deeper network's ability to learn more abstract features. Complementarily, \cite{shen2020pavement} embraced the Cascade R-CNN \cite{cai2018cascade}, which is an extension of Faster R-CNN, to achieve augmented road crack detection performance. 

Unlike the R-CNN series that necessitates a separate region proposal generation stage, the YOLO series generally partition the road image into a collection of grids, directly producing class probabilities and offset values for each bounding box ensconced within these grids to detect cracks efficiently \cite{mandal2018automated,nie2019pavement,tsuchiya2019method,du2021pavement,ma2022automatic,guo2022pavement,xiang2022improved,du2022improvement,xiang2023road,su2024mod}. For instance, \cite{mandal2018automated} trained an YOLOv2 \cite{redmon2017yolo9000} network with a frozen ResNet-101 backbone for road crack detection, performing well in detecting alligator cracks. Further advancements saw the deployment of YOLOv3 \cite{redmon2018yolov3} in \cite{nie2019pavement} \cite{tsuchiya2019method} \cite{du2021pavement} for road crack detection, which demonstrated superior processing speeds over SSD-based and RCNN-based methods. Recent advancements also include modifications \cite{guo2022pavement} \cite{xiang2022improved} to YOLOv5 \cite{glenn2020yolov5}, which incorporated attention mechanisms and vision Transformer blocks to the backbone of YOLOv5 and performed better than YOLOv4 \cite{bochkovskiy2020yolov4}, YOLOv5 and Cascade R-CNN for road crack detection. 
Nevertheless, object detection-based methods can only recognize road cracks at the instance level, and they are infeasible when pixel-level road crack detection results are desired.

\begin{table*}[!t]
\begin{threeparttable}
\caption{List of selected supervised single modal-based pixel-wise road crack detection methods}
\label{table_singlesemantic}  
\begin{tabular}{R{1cm}R{2.4cm}R{0.6cm}C{2.2cm}m{10cm}}
\toprule
Type & Reference & Year & Key method & Description\\
\midrule 
\multirow{11}{*}[-10em]{\makecell[c]{CNN-Based}} & 
 Liu \etal \cite{liu2019deepcrack} & 2019 & FCN & Deepcrack \cite{liu2019deepcrack} incorporated a side-output layer into FCN and adopted condition random fields and guided filtering to obtain accurate crack detection results. \\
 & Zou \etal \cite{zou2018deepcrack} & 2018 & SegNet & Zou \etal \cite{zou2018deepcrack} fused features from various scales of SegNet to learn hierarchical information for enhanced road crack detection performance. \\
 & Han \etal \cite{han2021crackw} & 2021 & U-Net & In \cite{han2021crackw}, skip-level round-trip sampling blocks were designed and embedded to U-Net, which can enhance the network's memory of transmitting low-level features in shallow layers for improved road crack detection results. \\
 & Sun \etal \cite{sun2024ductnet} & 2024 & UNet & Sun \etal \cite{sun2024ductnet} proposed DUCTNet to detect cracks in road images captured by a UAV, which combines the densely connected structure of UNet++ \cite{zhou2019unet++} with the nested structure of U2Net \cite{qin2020u2} for powerful feature fusion and feature capture capabilities. \\
 & Yang \etal \cite{yang2019feature} & 2019 & Deeplab & 
 Yang \etal \cite{yang2019feature} proposed FPHBN, introducing side networks into the feature pyramid to boost hierarchical feature learning for improved performance in road crack detection. \\
 & Xu \etal \cite{xu2022pixel} & 2022 & Deeplab & Xu \etal \cite{xu2022pixel} advanced the detection capabilities of Deeplabv3 \cite{chen2017rethinking} by implementing a resolution maintain flow and a Stacked Atrous Spatial Pyramid Pooling module, facilitating better feature fusion and spatial information extraction to road cracks. \\
 & Gao \etal \cite{gao2023synergizing} & 2023 & Deeplab & In \cite{gao2023synergizing}, low-level and high-level feature extractors employing ASPP modules were designed in the encoder to obtain sufficient global and local information from road cracks. Subsequently, skip-connections were adopted in the decoder to effectively fuse high-level and low-level feature maps from the encoder, resulting in enhanced road crack detection performance. \\
 & Zhang \etal \cite{zhang2023ecsnet} & 2023 & \makecell[c]{Small Kernel \\ Convolution} & Zhang \etal \cite{zhang2023ecsnet} introduced an Efficient Crack Segmentation Neural Network (ECSNet), specifically designed for rapid real-time road crack detection. This network employs small kernel convolutional layers and parallel max pooling techniques, aimed at reducing model parameters while swiftly extracting crack information. \\
 & Zhu \cite{zhu2024lightweight} & 2024 & \makecell[c]{Hybrid Attention} & Zhu \cite{zhu2024lightweight} proposed a lightweight encoder-decoder network known as RHACrackNet for road crack detection. By designing a novel hybrid attention block and adding residual blocks to the deep layers of the encoder, the network's feature extraction capabilities can be preserved, even with a reduced number of model parameters. \\
 \midrule
\multirow{6}{*}[-7em]{\makecell[c]{Transformer- \\ Based}} 
 & Guo \etal \cite{guo2023pavement} & 2023 & SwinTransformer & Guo \etal \cite{guo2023pavement} unified SwinTransformer \cite{liu2021swin} blocks with multi-layer perception layers for pixel-wise road crack detection, outperforming FCN and Deeplabv3+, particularly in handling challenging images with shadows, dense cracks, and leaves. \\
 & Kuang \etal  \cite{kuang2024universal} & 2024 & Segformer & Kuang \etal  \cite{kuang2024universal} introduced a visual crack prompt (VCP) mechanism, which guides Segformer \cite{xie2021segformer} to focus more intently on high-frequency features of road cracks, resulting in improved detection performance. \\
 & Xu \etal \cite{xu2022pavement} & 2022 & \makecell[c]{Transformer, \\ Convolution} & A locally enhanced Transformer network (LETNet) was introduced, employing Transformer blocks to model long-range dependencies while integrating convolution-based local enhancement modules to compensate for the loss of local fine-grained features. This design provides a semantically rich feature representation for road crack detection. \\
 & Chen \etal \cite{chen2022refined} & 2022 & \makecell[c]{Transformer, \\ Convolution} & Building on the foundation of LETNet, Chen \etal \cite{chen2022refined} proposed LECSFormer, which incorporated cross-shaped transformer structures and a token shuffle operation to enrich information interaction across different channels, thereby enhancing long-range modeling capabilities for improved detection performance. \\
 & Bai \etal \cite{bai2023dmf} & 2023 & \makecell[c]{Transformer, \\ Convolution} & To achieve stronger feature representations, Bai \etal \cite{bai2023dmf} developed a dual-encoding path that simultaneously captures global context features from Transformer blocks and local detail information from CNN blocks. Additionally, an interactive attention learning (IAL) strategy was implemented to effectively fuse global and local features, improving the detection of minute details in road cracks. \\
 & Tao \etal \cite{tao2023convolutional} & 2023 & \makecell[c]{Transformer, \\ Convolution} & Tao \etal
 \cite{tao2023convolutional} substituted conventional convolution layers with specially designed dilated residual blocks to learn higher-level and clearer local crack features, while also introducing a boundary awareness module aimed at learning the boundary information of road cracks to refine detection results. \\
\bottomrule
\end{tabular} 
\end{threeparttable}
\end{table*}

\begin{table*}[!t]
\begin{threeparttable}
\caption{List of selected supervised data fusion-based pixel-wise road crack detection methods}
\label{table_datafusionsemantic}  
\begin{tabular}{C{2.8cm}C{0.7cm}C{2cm}p{11
cm}}
\toprule
Reference & Year & Input & Description\\
\midrule 
 \makecell[c]{\hfill \\ Li \etal \cite{li2024cnn}} & \makecell[c]{\hfill \\ 2024} & \makecell[c]{\hfill \\ Gray image, \\ Depth maps} & {\vspace{-15pt} Li \etal \cite{li2024cnn} proposed a double-head U-Net (DHU-Net) incorporating channel attention module and spatial attention module to separately extract features from gray images and depth maps, achieving superior crack detection performance compared to U-Net, SegNet and Deeplabv3.} \\
\makecell[c]{\hfill \\ Jing \etal \cite{jing2024self}} & \makecell[c]{\hfill \\ 2024} & \makecell[c]{\hfill \\ Color image, \\ Depth maps} & {\vspace{-15pt} Jing \etal \cite{jing2024self} proposed CSF-Cracknet, which integrates color information from RGB images with structural details from depth maps by adaptively adjusting weights across image channels and spatial regions, which can be flexibly deployed at the front end of any semantic segmentation network to enhance road crack detection performance.} \\
\bottomrule
\end{tabular} 
\end{threeparttable}
\end{table*}
    
\subsection{Semantic Segmentation Networks}
\label{semantic}

As illustrated in Fig. \ref{fig_outline}, SOTA semantic segmentation networks for road crack detection can be broadly classified into two categories: (1) single-modal and (2) data-fusion approaches. Single-modal networks generally segment RGB images at the pixel-level with CNN-based and Transformer-based methods, whereas data-fusion networks integrate visual features from multiple types of vision sensor data to achieve a more comprehensive semantic understanding of the environment. A summary of the most representative methods is provided in Table \ref{table_singlesemantic} and Table \ref{table_datafusionsemantic}.

\subsubsection{Single modal-based}

CNN-based semantic segmentation methods designed for road crack detection are mainly developed from basic model architectures, including FCN \cite{long2015fully}, SegNet \cite{badrinarayanan2017segnet}, U-Net \cite{ronneberger2015u}, DeepLab \cite{chen2014semantic}, \etc. For instance, \cite{yang2018automatic} and \cite{dung2019autonomous} introduced FCN with VGG-19 and VGG-16 \cite{simonyan2014very} as the backbones to detect road cracks, respectively. \cite{choi2019sddnet} proposed SDDNET by introducing separable convolutions and dilated convolutions to FCN for real-time crack segmentation. In \cite{liu2019computer}, a U-Net model was trained to detect cracks in concrete road surface, showing better performance than two FCN-based methods \cite{yang2018automatic} \cite{dung2019autonomous}. \cite{sun2022dma} proposed DMA-Net, integrating a multi-scale attention module into the decoder of Deeplabv3+ \cite{chen2018encoder} to generate an attention mask and dynamically assign weights across different feature maps, thereby enhancing road crack detection results.
Moreover, to enhance the deployment of crack detection algorithms in practical applications by balancing detection performance with real-time processing requirements, several lightweight CNN-based approaches have been developed. For instance, in \cite{10539282}, edge extraction modules (EEM) based on traditional image processing methods, alongside parallel feature extractor modules (PFM), were devised to capture detailed feature information without resorting to very deep network architectures, thereby significantly reducing trainable parameters. 

While CNN-based crack detection algorithms demonstrate commendable performance in road crack detection, they inherently struggle with explicitly modeling long-range dependencies due to the localized nature of convolution operations. This limitation is particularly evident in scenarios involving elongated and narrow cracks, as well as in conditions where there is low contrast between the crack and the road surface. Transformers, with their sequence-to-sequence prediction capabilities and innate global self-attention mechanisms, have emerged as promising alternatives to address these challenges. For instance, \cite{liu2021crackformer} introduced CrackFormer, a SegNet-inspired encoder-decoder architecture incorporating self-attention modules to capture extensive contextual information, alongside scaling-attention modules to suppress non-semantic features and enhance semantic ones. Further advancements by \cite{liu2023crackformer} enhanced this architecture, employing improved Transformer blocks with local self-attention layers, local feed-forward layers, and skip connections, thereby facilitating more efficient contextual information capture. 

However, the exclusive use of Transformer blocks may lead to limited localization abilities due to a deficiency in capturing low-level details. Therefore, the advanced pixel-wise road crack detection algorithms \cite{fang2022external,xiao2023pavement,xu2022pavement,shamsabadi2022vision,wang2022automatic,chen2022refined,tao2023convolutional,bai2023dmf} tend to adopt a hybrid CNN-Transformer architecture, enabling the integration of both detailed hierarchical spatial information and global long-range contextual information, thereby facilitating precise detection results. For instance, \cite{shamsabadi2022vision} introduced TransUNet \cite{chen2021transunet} for pixel-wise road crack detection, which combines Transformer blocks as a strong encoder with U-Net to enhance fine details by restoring localized spatial information. Further advancing the robustness of TransUNet, \cite{fang2022external} developed an external attention block to exploit the dependencies of crack regions across different images. Experimental results indicated  \cite{shamsabadi2022vision} \cite{fang2022external} can effectively mitigate interferences from shadows and noises.  

\subsubsection{Data-fusion-based}

The fusion of diverse vision sensor data has emerged as a prominent topic in the field of computer vision and robotics. Specially, RGB and gray images provide detailed insights into road surface texture, while other types of road data can complement this by providing crucial 3D information, polarization light information, thermal distribution, \etc, especially for detecting cracks not easily visible in 2D images. For instance, in \cite{guan2021automated}, RGB images and depth maps were combined channel-wise to train a lightweight U-Net, resulting in improved road crack detection performance compared to using a single type of road data. \cite{zhou2021crack} compared two networks designed to fuse RGB images and range images for pixel-wise road crack detection. Experimental results demonstrated that the network performed separate feature extraction on the two types of road data and then fused high-level features outperformed the network directly utilizing the combined image data. However, data-fusion-based road crack detection methods face significant challenges. The complexity and computational demands of processing and integrating these distinct data types require advanced algorithms and considerable processing power, resulting in higher hardware and software costs. 

\begin{table*}[!t]
\begin{threeparttable}
\caption{List of selected label-efficient deep learning-based road crack detection methods}
\label{table_underannotated}  
\begin{tabular}{R{1cm}R{2.4cm}R{0.6cm}C{2.8cm}m{9.1cm}}
\toprule
Type & Reference & Year & Key method & Description\\
\midrule
\multirow{2}{*}[-0.5em]{\makecell[c]{Unsupervised}} 
& Yu \etal \cite{yu2020unsupervised} & 2020 & \makecell[c]{ Adversarial \\ Image-to-Frequency \\ Transform (AIFT)} & In Yu \etal \cite{yu2020unsupervised}, an AIFT architecture was developed. This method uses only healthy road images to derive a transformation model between the image and frequency domains. Road cracks are then detected by comparing the given and generated damaged road images within each domain.  \\
 & Ma \etal \cite{ma2024up} & 2024 & \makecell[c]{Generative Adversarial \\ Network (GAN)} & Ma \etal \cite{ma2024up} proposed UP-CrackNet, utilizing multi-scale square masks to corrupt healthy road images, and training a GAN restore the corrupted regions by leveraging the semantic context learned from the surrounding uncorrupted regions. In the testing phase, an error map is created by calculating the difference between the input and restored images, enabling pixel-wise crack detection. \\
\midrule
\multirow{2}{*}[-0.5em]{\makecell[c]{Semi- \\ supervised}} 
 & Wang \etal \cite{wang2021semi} & 2021 & \makecell[c]{Teacher-Student \\ Architecture} & Wang \etal \cite{wang2021semi} introduced a teacher-student architecture. By enforcing the output consistency between the two models under added noise, additional training signals can be extracted from unlabeled data, thereby enhancing the student model's representation for improved road crack detection performance. \\
 & Liu \etal \cite{liu2024semi} & 2024 & \makecell[c]{Cross-Consistency \\ Training (CCT)} & In Liu \etal \cite{liu2024semi}, CCT \cite{ouali2020semi} was implemented for semi-supervised road crack detection. This approach leveraged unlabeled data to improve the performance of the main segmentation network by enforcing consistency between the main decoder and auxiliary decoders, where the input of auxiliary decoders are perturbed versions of the main encoder's output.  \\
 \midrule
\multirow{2}{*}[-0.5em]{\makecell[c]{Weakly- \\ supervised}} 
 & Al \etal \cite{al2023weakly} & 2023 & \makecell[c]{Multi-scale \\ CAM Generation} & In Al \etal \cite{al2023weakly}, the contrast limited adaptive histogram equalization (CLAHE) \cite{pizer1990contrast} technique was adopted to mitigate the detrimental effects of uneven illumination on input images, and a multi-scale CAM generation strategy was proposed to produce higher-quality pseudo-labels. \\
 & Zhang \etal \cite{zhang2022investigation} & 2022 & \makecell[c]{Region Growing, \\ GrabCut algorithm} & \cite{zhang2022investigation} utilized a region growing algorithm and a GrabCut algorithm to generate pixel-wise pseudo-labels from the bounding box-level detection results. \\
\bottomrule
\end{tabular} 
\end{threeparttable}
\end{table*}

\section{Label-efficient Deep learning-based Road Crack Detection Methods}
\label{sec.under}

Although the semantic segmentation algorithms discussed in Section \ref{semantic} demonstrate significant potential for accurately detecting road cracks at the pixel level, their training necessitates extensive datasets with meticulously detailed human annotations, resulting in considerable labor and time costs. To mitigate this challenge, researchers have developed label-efficient deep learning-based methods for road crack detection, which can be classified into unsupervised, semi-supervised, and weakly-supervised. The selected methods are summarized in Table \ref{table_underannotated}.

Unsupervised road crack detection methods can be divided into image space-based and feature space-based. The core mechanism involves training networks using only healthy road images to develop strong restoration and discrimination capabilities in the image or feature space. During the testing phase, when provided with a damaged road image, the trained network can restore undamaged regions but struggle with reconstructing crack regions. Consequently, the discrepancy between the input damaged image and the restored image can be utilized to generate pixel-wise crack detection results. However, unsupervised road crack detection methods tend to encounter challenges in accurately handling fine details and small anomalies. They may unexpectedly restore tiny cracks to their original appearance, leading to the missed detections, or fail to restore unseen disturbances (watermark digits, shadows, \etc), resulting in false detections. Therefore, semi-supervised and weakly-supervised methods, which utilize limited or coarse label information, offer more robust alternatives for road crack detection. 

The existing semi-supervised road crack detection algorithms typically use consistency training, combining a small set of labeled road images along with a larger set of unlabeled ones to effectively train a segmentation network to detect cracks. Cross-consistency training enforces prediction invariance under perturbations, enhancing the model's robustness to data distribution changes and improving the detection performance compared to only using supervised learning with limited labeled data. 

The typical workflow for weakly-supervised road crack detection algorithms involves three key steps: (1) training a model using coarse labels, such as image-level \cite{dong2020patch,konig2022weakly,al2023weakly} or bounding box-level \cite{zhang2022investigation} annotations, to generate Class Activation Maps (CAMs) \cite{zhou2016learning} or approximate crack locations from neural layers; (2) refining these CAMs or instance-level detection results into pixel-wise pseudo-labels; and (3) using these pseudo-labels to train a semantic segmentation model for pixel-wise road crack detection. For instance, \cite{dong2020patch} employed the discriminative localization technique to derive CAMs of each image patch and utilized the DenseCRF method \cite{krahenbuhl2011efficient} to generate pixel-wise pseudo-labels to train a SegNet \cite{badrinarayanan2017segnet} model for road crack detection. \cite{konig2022weakly} further enhanced the quality of pseudo-labels derived from CAMs by integrating them with segmentation maps obtained through multi-Otsu's thresholding \cite{otsu1975threshold}. 

\begin{table*}[!t]
\begin{threeparttable}
\caption{Selected public image-based datasets for road crack detection}
\label{table}  
\begin{tabular}{R{1cm}R{2cm}R{1cm}R{1.8cm}m{10.4cm}}
\toprule
Type & Dataset & Amount & Resolution & Description\\
\midrule
\multirow{3}{*}[-0.5em]{Image-level} 
 & \makecell[r]{SDNET2018 \\ \cite{maguire2018sdnet2018}} & 56,000 & $256\times256$ & SDNET2018 dataset\footnotemark[6] was captured using a 16 MP Nikon digital camera at Utah State University, encompassing images of cracks found in concrete bridge decks, roads, and walls, with widths ranging from 0.66 mm to 25 mm. \\
 & \makecell[r]{CQU-BPDD \\ \cite{tang2021iteratively}} & 60,059 &  $1200\times900$ &  CQU-BPDD dataset\footnotemark[7] was captured by the onboard cameras of a specialized road inspection vehicle operating in various regions of southern China, including seven types of bituminous road cracks, such as transverse, alligator, and longitudinal cracks.\\
\midrule
\multirow{6}{*}[-0.1em]{\makecell{\makecell{Bounding \\ Box-level}}} 
 & \makecell[r]{RDD2018 \\ \cite{maeda2018road}} & 9,053 & $600\times600$ & RDD2018 dataset\footnotemark[8] was collected using a smartphone mounted on a vehicle, comprising $9,053$ road damage images with $15,435$ instances of road surface damage. The dataset covers seven municipalities in Japan, exhibiting diverse regional characteristics. 
 \\
 & \makecell[r]{RDD2020 \\ \cite{arya2021rdd2020}} & 26,336 &    \makecell[r]{$600\times600$ \\ $720\times720$} & RDD2020 dataset\footnotemark[9] contains $10,506$ images from Japan, $2,829$ images from Chile and $7,706$ images from India. over $31,000$ instances of road damage across four categories are collected: longitudinal cracks, transverse cracks, alligator cracks, and potholes. \\
 & \makecell[r]{RDD2022 \\ \cite{arya2024rdd2022}} & 47,420 
 & \makecell[r]{$512\times512$ \\ $600\times600$ \\ $720\times720$}
 & RDD2022 dataset\footnotemark[10] extended \cite{arya2021rdd2020} by adding road images from Norway, the United States, and China, while maintaining the same four categories of road damage. Vehicle-based systems, including motorcycles, UAVs, and cars, are used as the acquisition equipment.
 \\
\midrule
\multirow{14}{*}{Pixel-level} 
 & \makecell[r]{CrackTree260 \\ \cite{zou2012cracktree}} & 260 & $800\times600$ & CrackTree260 dataset\footnotemark[11] contains four types of road cracks: alligator, longitudinal, transverse and multifurcate. Some images have shadows, zebra crossing markings, and oil spots.
 \\
 & CFD \cite{shi2016automatic} & 118 & $480\times320$ & CrackForest dataset\footnotemark[12] (CFD) contains $118$ images of road cracks with widths ranging from $1$ to $3$ $mm$, presenting a range of illumination conditions, shadows, and stains.
 \\
 & AigleRN \cite{amhaz2016automatic} & 38 & $311\times462$ & AigleRN dataset\footnotemark[13] contains four types of cracks with pixel-level annotations: alligator, longitudinal, transverse, and block. All images are annotated with pixel-level labels and pre-processed to mitigate the influence of non-uniform lighting conditions.
 \\
 & \makecell[r]{CRKWH100 \\ \cite{zou2018deepcrack}} & 100 & $512\times512$ & 
 CRKWH100 dataset\footnotemark[14] includes $100$ road images captured using a line-array camera under consistent lighting conditions, achieving a ground sampling distance of $1$ millimeter. 
 \\
 & \makecell[r]{CrackLS315 \\ \cite{zou2018deepcrack}} & 315 & $512\times512$ & CrackLS315 dataset\footnotemark[14] was also collected using a line-array camera, at the same ground sampling distance as CRKWH100 dataset. \\
 & Stone331 \cite{zou2018deepcrack} & 331 & $1024\times1024$ & Stone331 dataset\footnotemark[14] comprises $331$ images of stone surfaces with pixel-level annotations, acquired using an area-array camera under visible-light illumination during the stone cutting process. For each image, a mask is generated to delineate the area of the stone surface.
 \\
 & CCSD \cite{ozgenel2019concrete} & 458 & $608\times608$ & 
 Concrete crack segmentation dataset\footnotemark[15] (CCSD) was collected from various locations at the Middle East Technical University, containing $458$ images on the concrete road surface, accompanied by the corresponding alpha maps indicating the presence of cracks.
 \\
 & \makecell[r]{ShadowCrack \\ \cite{fan2023pavement}} & 210 & $480\times480$ &  ShadowCrack dataset\footnotemark[16] was captured using an iPhone XR at a height of 1 to 1.2 meters in Beijing and Changchun, China. It includes shadows cast by various traffic objects and urban features, such as vehicles, pedestrians, trees, and buildings. 
 \\
 & \makecell[r]{CrackNJ156 \\ \cite{xu2022pavement}} & 156 & $1734\times1734$ & CrackNJ156 dataset\footnotemark[17] comprises 156 pavement surface (asphalt, concrete, terrazzo, \etc) images, containing cracks under diverse weather, season, and lighting conditions.
 \\
 & \makecell[r]{NHA12D \\ \cite{huang2022nha12d}} & 80 & $1920\times1080$ &
 NHA12D dataset\footnotemark[18] contains $40$ images of concrete road surface and $40$ images of asphalt road surface. These images were captured by National Highway's digital survey vehicles on the A12 network in the UK. \\
 & \makecell[r]{CrackSC \\ \cite{guo2023pavement}} & 197 & $320\times480$ & 
 CrackSC dataset\footnotemark[19] was collected using an iPhone 8 around Enoree Avenue in Columbia, South Carolina, United States. It focuses on the heavy shadows and dense cracks typically found on local roads, having tree shadows, fallen leaves, and moss.
\\
\bottomrule
\end{tabular} 
\end{threeparttable}
\end{table*}

\begin{table*}[t!]
	\renewcommand{\arraystretch}{1}
	\settablefont
	\caption{Quantitative experimental results of pixel-wise crack detection performance on the proposed UDTIRI-Crack dataset}
	\centering
    \begin{threeparttable}
	\begin{tabular}{C{1.1cm}C{1.5cm}C{2.5cm}C{1.6cm}C{1.4cm}C{1.6cm}C{1.6cm}C{1.3cm}C{1.3cm}}
		\toprule
		\multicolumn{1}{c}{Mode} & \multicolumn{1}{c}{Training Strategy} & {Methods} & Precision~($\%$)$\uparrow$ & Recall~($\%$)$\uparrow$ & Accuracy~($\%$)$\uparrow$ & F1-Score~($\%$)$\uparrow$ & IoU~($\%$)$\uparrow$ & AIoU~($\%$)$\uparrow$ \\ \midrule
		\multirow{10}{*}{CNN} & \multirow{4}{*}{General Supervised}
        &{BiseNet \cite{yu2018bisenet}} & 72.621 & 52.328 & 98.299 & 60.826 & 43.705 & 70.991 \\
        &&{PSPNet \cite{zhao2017pyramid}} & 73.162 & 53.570 & 98.332 & 61.852 & 44.772 & 71.541 \\
        &&{LEDNet \cite{wang2019lednet}}  & 73.973 & 47.936 & 98.260 & 58.174 & 41.018 & 69.628 \\
        &&{DeepLabv3+ \cite{chen2018encoder}}  & 73.965 & 54.047 & 98.360 & 62.457 & 45.409 & 71.873 \\
        \cmidrule(lr){2-9}
        &\multirow{6}{*}{\makecell[c]{Crack Detection- \\ Specific Supervised}}
        &{ECSNet \cite{zhang2023ecsnet}}  & \textbf{77.678} & 53.584 & 98.440 & 63.420 & 46.434 & 72.426 \\
        &&{Deepcrack18 \cite{zou2018deepcrack}} & 74.736 & 58.702 & 98.457 & 65.756 & 48.982 & 73.708 \\
        &&{Deepcrack19 \cite{liu2019deepcrack}} & 72.831 & 57.795 & 98.390 & 64.448 & 47.544 & 72.956\\
        &&{SCCDNet \cite{li2021sccdnet}} & 71.294 & 58.371 & 98.356 & 64.189 & 47.263 & 72.797 \\
        &&{Crack-Att \cite{xu2023crack}} & 68.790 & 66.447 & 98.392 & 67.598 & 51.055 & 74.710 \\
        &&{CDLN \cite{manjunatha2024crackdenselinknet}} & 56.947 & \textbf{82.918} & 97.986 & 67.522 & 50.968 & 74.456\\  
        &&{SegDecNet++ \cite{tabernik2023automated}} & 66.185 & 66.890 & 98.302 & 66.536 & 49.853 & 74.063\\  
		\midrule
		\multirow{9}{*}{Transformer} & \multirow{7}{*}{General Supervised}
        & LM-Net\cite{lu2024lm} & 77.409 & 62.434 & 98.592 & 69.120 & 52.811 & 75.690 \\
       &&{SwinTransformer \cite{liu2021swin}} & 75.065 & 65.704 & 98.583 & 70.073 & 53.933 & 76.246 \\
        &&{SegFormer \cite{xie2021segformer}} & 58.973 & 67.097 & 97.991 & 62.773 & 45.744 & 71.850\\
        &&{TransUnet\cite{chen2021transunet}} & 68.121 & 68.108 & 98.390 & 68.115 & 51.647 & 75.005\\
        &&{SCTNet \cite{xu2024sctnet}} & 73.930 & 59.902 & 98.455 & 66.181 & 49.456 & 73.943 \\
        &&{AFFormer \cite{dong2023head}} & 76.920 & 63.258 & 98.593 & 69.423 & 53.167 & 75.869 \\
        \cmidrule(lr){2-9}
        &\multirow{2}{*}{\makecell[c]{Crack Detection- \\ Specific Supervised}}
        &{LECSFormer \cite{chen2022refined}} & 74.711 & 65.341 & 98.567 & 69.712 & 53.506 & 76.025\\
        &&{CT-crackseg \cite{tao2023convolutional}} & 75.019 & 66.694 & \textbf{98.599} & \textbf{70.612} & \textbf{54.573} & \textbf{76.574} \\
		\bottomrule
	\end{tabular}
    \end{threeparttable}
	\label{table_UDTIRI-Crack}
\end{table*}

\section{Benchmark Analysis} 
\label{sec.experiment}

\subsection{Image-Based Road Crack Datasets}

Image-based road crack datasets serve as critical benchmarks, facilitating the rigorous evaluation and comparison of advanced road crack detection algorithms. They can be categorized into three types according to their annotation methods, including image-level, bounding box-level, and pixel-level. For instance, 
the Concrete Crack Images for Classification (CCI4C) dataset\footnote{\href{CCI4C}{https://data.mendeley.com/datasets/5y9wdsg2zt/2}} \cite{ozgenel2019concrete} was collected from various locations on the Middle East Technical University campus. $20,000$ images with a resolution of 227 $\times$ 227 pixels are categorized into two classes (crack and non-crack) with image-level labels. Tailored for object detection task, the German Asphalt Pavement Distress (GAPs) dataset\footnote{\href{GAPsv1}{http://www.tu-ilmenau.de/neurob/data-sets-code/gaps/}} \cite{eisenbach2017get} was established, consisting of $1,969$ images with bounding box-level labels of road cracks, where the image resolution is $1920\times1080$ pixels. \cite{stricker2019improving} further extended GAPs to $2,468$ images and improved the quality of labels using bounding boxes with minimal overlap. To further provide precise location of cracks, the Crack500 \cite{yang2019feature} dataset\footnote{\href{crack500}{https://github.com/fyangneil/pavement-crack-detection}} was collected using a smartphone at the main campus of Temple University and human-annotated with pixel-level labels, containing $500$ images of pavement crack of size $2000 \times 1500$ pixels. It contains four types of cracks (alligator, longitudinal, transverse, and multifurcate), presenting substantial hurdles for realistic crack detection, due to the presence of obstructions, shadows, and diverse lighting scenarios. Similarly, the DeepCrack \cite{liu2019deepcrack} dataset\footnote{\href{deepcrack}{https://github.com/yhlleo/DeepCrack}} comprises $537$ concrete and asphalt surface images with cracks of varying scales and scenes. Each image is captured at a resolution of $544\times384$ pixels and meticulously human-annotated at the pixel level. As summarized in TABLE \ref{table}, selected public road crack datasets are detailed with annotation types, names, data volume, image resolutions, and descriptions. 

\footnotetext[5]{https://www.kaggle.com/datasets/jefffffffsong/udtiri-crack}

Given the variability in image quality and annotation accuracy across existing pixel-level annotated road crack datasets, as well as the lack of a dataset that ensures sufficient diversity and balance in crack types and styles, we curated a high-quality dataset named UDTIRI-Crack\footnotemark[5]. This dataset consists of $2,500$ images (resolution: $320\times320$ pixels) sourced from seven public datasets: Crack500 \cite{yang2019feature}, CrackLS315 \cite{zou2018deepcrack}, CrackSC \cite{guo2023pavement}, CrackTree260 \cite{zou2012cracktree}, CRKWH100 \cite{zou2018deepcrack}, DeepCrack537 \cite{liu2019deepcrack}, and ShadowCrack \cite{fan2023pavement}. UDTIRI-Crack includes five common types of road cracks: alligator, longitudinal, transverse, multifurcate, and pit cracks. It also spans a variety of pavement materials, such as concrete and asphalt, and covers diverse scenes and lighting conditions. Additionally, the dataset incorporates various noise factors, including shadows from traffic objects, zebra crossing markings, oil spots, obstructions, fallen leaves, and moss. The UDTIRI-Crack dataset is divided into $1,500$ images for training, $400$ for validation, and $600$ for testing, which has been promoted as the first extensive online benchmark for road crack detection.

\footnotetext[6]{https://digitalcommons.usu.edu/all\_datasets/48/}
\footnotetext[7]{https://whtang.cn/CQU-BPDD/}
\footnotetext[8]{https://github.com/sekilab/RoadDamageDetector/}
\footnotetext[9]{https://data.mendeley.com/datasets/5ty2wb6gvg/1}
\footnotetext[10]{https://figshare.com/articles/dataset/RDD2022}
\footnotetext[11]{https://1drv.ms/f/s!AittnGm6vRKLyiQUk3ViLu8L9Wzb}
\footnotetext[12]{https://github.com/cuilimeng/CrackForest-dataset}
\footnotetext[13]{https://github.com/Sutadasuto/uvgg19\_crack\_detection}
\footnotetext[14]{https://1drv.ms/f/s!AittnGm6vRKLtylBkxVXw5arGn6R}
\footnotetext[15]{https://data.mendeley.com/datasets/jwsn7tfbrp/1}
\footnotetext[16]{https://github.com/fanlili666/shadow-crack-dataset}
\footnotetext[17]{https://zenodo.org/records/6526409}
\footnotetext[18]{https://github.com/ZheningHuang?tab=repositories}
\footnotetext[19]{https://github.com/jonguo111/Transformer-Crack}

\begin{figure*}[!t]
		\includegraphics[width=\textwidth]{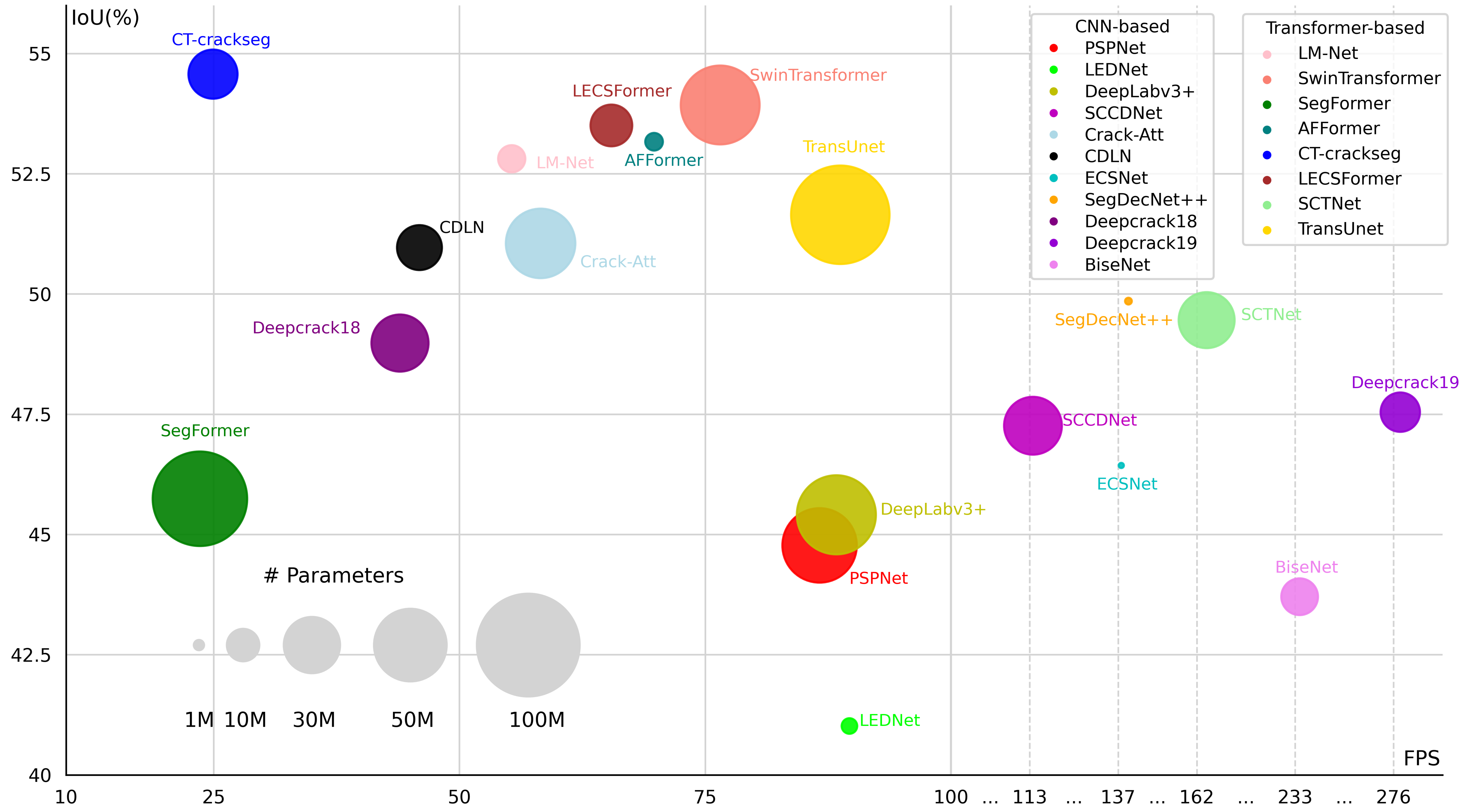}
        \captionsetup{font={small}}
		\caption{Comparison results of eleven CNN-based and eight Transformer-based supervised semantic segmentation methods in terms of detection performance (IoU), resource consumption (model parameters) and computational complexity (FPS).}
		\label{fig_UDTIRI-Crack}
\end{figure*}

\subsection{Implementation Details}
To evaluate the detection performance of the existing semantic segmentation methods for road crack detection, we conduct experiments using ten public general-purpose supervised methods and nine crack detection-specific supervised methods on the proposed UDTIRI-Crack dataset. Furthermore, to assess the generalizability of these methods under shadow conditions, diverse weather and season conditions, we extract $100$ and $200$ image patches (resolution:$320\times320$ pixels) from the AigleRN \cite{amhaz2016automatic} and CrackNJ156 \cite{xu2022pavement} datasets, respectively. These image patches are used as the test data to evaluate the performance of models trained on the UDTIRI-Crack dataset.  

All experiments are conducted on a single NVIDIA RTX 3090 GPU, with models trained for $200$ epochs using a batch size of $4$. The learning rate was initialized at $0.001$ and adjusted dynamically following the poly learning rate scheduling strategy. The Adam optimizer, configured with $\beta_1=0.5$ and $\beta_2=0.999$, is employed to optimize the networks. To holistically and fairly evaluate the detection performance of the compared methods, we utilize precision, recall, accuracy, intersection over union (IoU), F1-Score, and average intersection over union (AIoU) as quantitative metrics, where AIoU represents the average IoU of crack and background categories. In addition, to comprehensively assess the compared methods' resource consumption and computational complexity, we further employ model parameters and frames per second (FPS) as quantitative metrics.  

\subsection{Comparison Results}

\begin{table*}[t!]
	\renewcommand{\arraystretch}{1}
	\settablefont
	\caption{Quantitative experimental results of pixel-wise crack detection performance on the AigleRN dataset \cite{amhaz2016automatic}}
	\centering
    \begin{threeparttable}
	\begin{tabular}{C{1.1cm}C{1.5cm}C{2.5cm}C{1.6cm}C{1.4cm}C{1.6cm}C{1.6cm}C{1.3cm}C{1.3cm}}
		\toprule
		\multicolumn{1}{c}{Mode} & \multicolumn{1}{c}{Training Strategy} & {Methods} & Precision~($\%$)$\uparrow$ & Recall~($\%$)$\uparrow$ & Accuracy~($\%$)$\uparrow$ & F1-Score~($\%$)$\uparrow$ & IoU~($\%$)$\uparrow$ & AIoU~($\%$)$\uparrow$ \\ \midrule
		\multirow{10}{*}{CNN} & \multirow{4}{*}{General Supervised} 
        & {BiseNet \cite{yu2018bisenet}} & 51.930 & 27.181 & 99.016 & 35.685 & 21.717 & 60.365 \\
		& & {PSPNet \cite{zhao2017pyramid}} & 49.797 & 30.787 & 98.994 & 38.050 & 23.495 & 61.243 \\
		& & {LEDNet \cite{wang2019lednet}} & 46.214 & 26.389 & 98.953 & 33.594 & 20.188 & 59.569 \\
		& & {DeepLabv3+ \cite{chen2018encoder}} & 59.131 & 44.943 & 99.135 & 51.070 & 34.291 & 66.711 \\
		\cmidrule(lr){2-9}
		& \multirow{6}{*}{\makecell[c]{Crack Detection- \\ Specific Supervised}} 
        & {ECSNet \cite{zhang2023ecsnet}} & 67.145 & 46.927 & 99.237 & 55.244 & 38.164 & 68.698 \\
		& & {Deepcrack18 \cite{zou2018deepcrack}} & 66.291 & 70.187 & 99.342 & 68.183 & 51.726 & 75.532 \\
		& & {Deepcrack19 \cite{liu2019deepcrack}} & 71.502 & 59.303 & 99.354 & 64.834 & 47.966 & 73.658 \\
		& & {SCCDNet \cite{li2021sccdnet}} & 80.847 & 56.591 & 99.430 & 66.579 & 49.901 & 74.664 \\
		& & {Crack-Att \cite{xu2023crack}} & 61.637 & 74.493 & 99.278 & 67.458 & 50.895 & 75.084 \\
		& & {CDLN \cite{manjunatha2024crackdenselinknet}} & 52.604 & \textbf{87.651} & 99.083 & 65.749 & 48.974 & 74.025 \\  
		& & {SegDecNet++ \cite{tabernik2023automated}} & 64.358 & 68.533 & 99.303 & 66.380 & 49.678 & 74.488 \\  
		\midrule
		\multirow{9}{*}{Transformer} & \multirow{7}{*}{General Supervised}  & LM-Net\cite{lu2024lm} & 81.923 & 65.292 & \textbf{99.507} & \textbf{72.668} & \textbf{57.070} & \textbf{78.287 }
        \\&& {SwinTransformer \cite{liu2021swin}} & 81.637 & 16.143 & 99.122 & 26.956 & 15.577 & 57.349 \\
		& & {SegFormer \cite{xie2021segformer}} & 59.104 & 67.741 & 99.206 & 63.129 & 46.123 & 72.661 \\
		& & {TransUnet\cite{chen2021transunet}} & 59.607 & 25.330 & 99.078 & 35.552 & 21.619 & 60.347 \\
		& & {SCTNet \cite{xu2024sctnet}}  & 67.085 & 15.765 & 99.077 & 25.530 & 14.633 & 56.854 \\
		& & {AFFormer \cite{dong2023head}} & 73.455 & 20.238 & 99.126 & 31.733 & 18.859 & 58.991 \\
		\cmidrule(lr){2-9}
		& \multirow{2}{*}{\makecell[c]{Crack Detection- \\ Specific Supervised}}& {LECSFormer \cite{chen2022refined}}  & 63.742 & 66.123 & 99.282 & 64.910 & 48.050 & 73.664\\
		& & {CT-crackseg \cite{tao2023convolutional}} & \textbf{92.393} & 45.012 & 99.411 & 60.533 & 43.403 & 71.406\\
		\bottomrule
	\end{tabular}
    \end{threeparttable}
	\label{table_aiglern}
\end{table*}

\begin{table*}[t!]
	\renewcommand{\arraystretch}{1}
	\settablefont
	\caption{Quantitative experimental results of pixel-wise crack detection performance on the CrackNJ156 dataset \cite{xu2022pavement}}
	\centering
    \begin{threeparttable}
	\begin{tabular}{C{1.1cm}C{1.5cm}C{2.5cm}C{1.6cm}C{1.4cm}C{1.6cm}C{1.6cm}C{1.3cm}C{1.3cm}}
		\toprule
		\multicolumn{1}{c}{Mode} & \multicolumn{1}{c}{Training Strategy} & {Methods} & Precision~($\%$)$\uparrow$ & Recall~($\%$)$\uparrow$ & Accuracy~($\%$)$\uparrow$ & F1-Score~($\%$)$\uparrow$ & IoU~($\%$)$\uparrow$ & AIoU~($\%$)$\uparrow$ \\ \midrule
		\multirow{10}{*}{CNN} & \multirow{4}{*}{General Supervised} &{BiseNet \cite{yu2018bisenet}} & 31.812 & 33.848 & 96.008 & 32.798 & 19.616 & 57.792 \\
		&&{PSPNet \cite{zhao2017pyramid}} & 29.165 & 38.699 & 95.531 & 33.263 & 19.949 & 57.715 \\
        &&{LEDNet \cite{wang2019lednet}}  & 42.596 & 30.108 & 96.821 & 35.279 & 21.418 & 59.105 \\
        &&{DeepLabv3+ \cite{chen2018encoder}}  & 37.466 & 31.897 & 96.508 & 34.458 & 20.815 & 58.645 \\
        \cmidrule(lr){2-9}
        &\multirow{6}{*}{\makecell[c]{Crack Detection- \\ Specific Supervised}}
        &{ECSNet \cite{zhang2023ecsnet}}  & 38.797 & 36.410 & 96.517 & 37.566 & 23.127 & 59.803 \\
        &&{Deepcrack18 \cite{zou2018deepcrack}} &32.269
        & 31.041 & 96.140 & 31.643 & 18.795 & 57.450 \\
        &&{Deepcrack19 \cite{liu2019deepcrack}} & 31.982 & 35.369 & 95.975 & 33.590 & 20.185 & 58.060\\
        &&{SCCDNet \cite{li2021sccdnet}} & 21.522 & 39.264 & 94.132 & 27.803 & 16.146 & 55.106 \\
        &&{Crack-Att \cite{xu2023crack}} & 42.843 & 40.439 & 96.733 & 41.606 & 26.267 & 61.481 \\
        &&{CDLN \cite{manjunatha2024crackdenselinknet}} & 18.274 & \textbf{59.664} & 91.160 & 27.979 & 16.265 & 53.635\\  
        &&{SegDecNet++ \cite{tabernik2023automated}} & 34.843 & 49.244 & 95.889 & 40.810 & 25.636 & 60.733\\  
		\midrule
		\multirow{9}{*}{Transformer} & \multirow{7}{*}{General Supervised}
         & LM-Net\cite{lu2024lm} & 38.330 & 39.892 & 96.423 & 39.095 & 24.297 & 60.339 \\
        &&{SwinTransformer \cite{liu2021swin}} & 32.769 & 53.267 & 95.510 & 40.576 & 25.452 & 60.446 \\
        &&{SegFormer \cite{xie2021segformer}} & 23.930 & 54.829 & 93.684 & 33.318 & 19.989 & 56.786 \\
        &&{TransUnet\cite{chen2021transunet}} & 36.703 & 54.350 & 95.989 & 43.816 & 28.054 & 61.990 \\
        &&{SCTNet \cite{xu2024sctnet}} & 36.568 & 45.026 & 96.170 & 40.359 & 25.281 & 60.700 \\
        &&{AFFormer \cite{dong2023head}} & \textbf{45.281} & 43.354 & \textbf{96.862} & 44.296 & 28.449 & \textbf{62.636} \\
        \cmidrule(lr){2-9}
        &\multirow{2}{*}{\makecell[c]{Crack Detection- \\ Specific Supervised}}
        &{LECSFormer \cite{chen2022refined}}  & 36.418 & 44.338 & 96.170 & 39.990 & 24.992 & 60.557 \\
        &&{CT-crackseg \cite{tao2023convolutional}} & 42.585 & 46.632 & 96.655 & \textbf{44.517} & \textbf{28.631} & 62.620 \\
		\bottomrule
	\end{tabular}
    \end{threeparttable}
	\label{table_cracknj}
\end{table*}

\begin{figure*}[!t]
	\begin{center}
		\centering
		\includegraphics[width=0.98\textwidth]{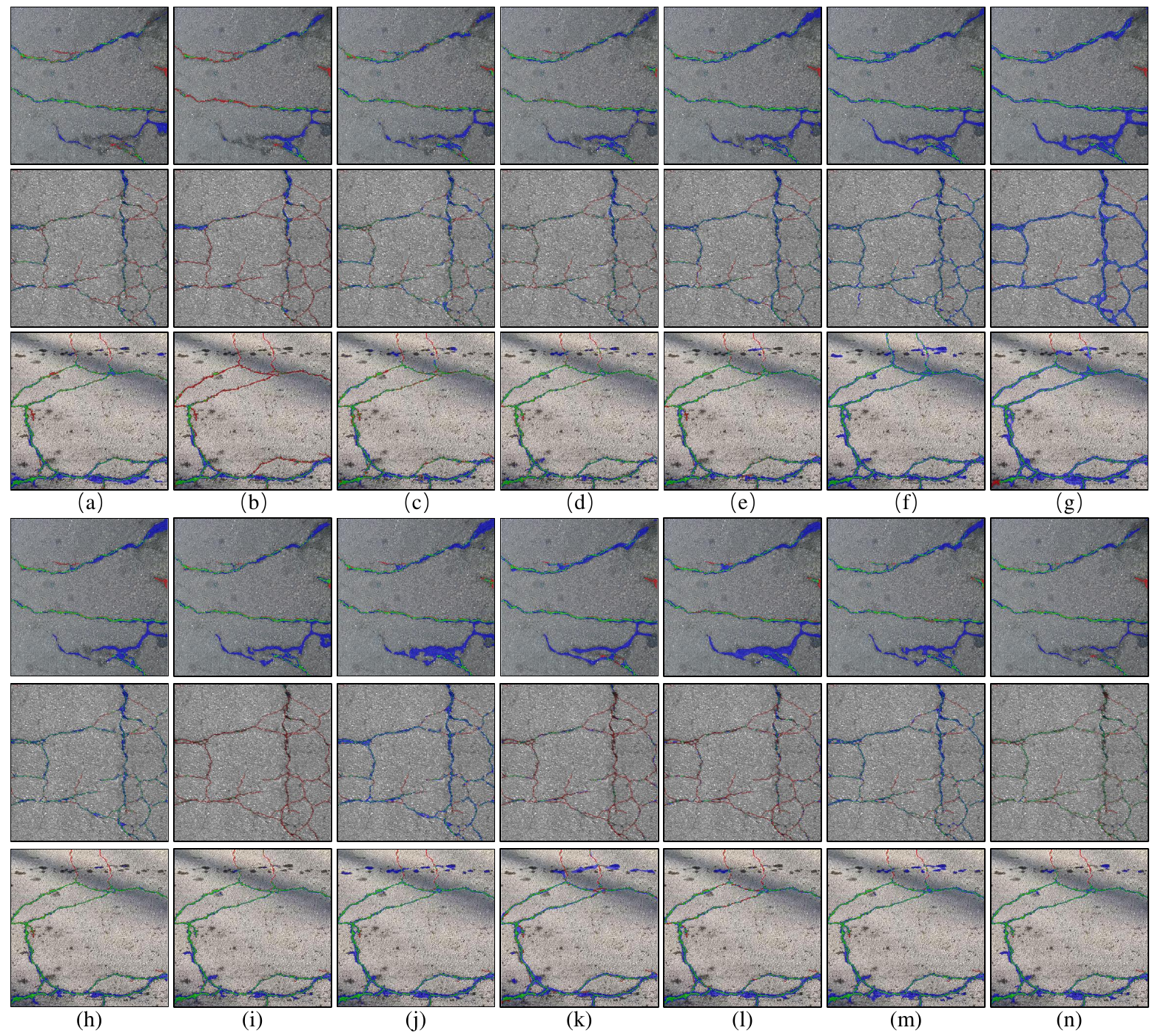}
		\centering
        \captionsetup{font={small}}
		\caption{Examples of experimental results of the compared methods on the proposed UDTIRI-Crack dataset: (a)  ECSNet \cite{zhang2023ecsnet}; (b) Deepcrack18 \cite{zou2018deepcrack}; (c) Deepcrack19 \cite{liu2019deepcrack}; (d) SCCDNet \cite{li2021sccdnet}; (e) Crack-Att \cite{xu2023crack}; (f) CDLN \cite{manjunatha2024crackdenselinknet}; (g) SegDecNet++ \cite{tabernik2023automated}; (h) LM-Net \cite{lu2024lm}
        (i) SwinTransformer \cite{liu2021swin};  (j) SegFormer \cite{xie2021segformer}; (k)
TransUnet\cite{chen2021transunet}; (l) SCTNet \cite{xu2024sctnet}; (m) LECSFormer \cite{chen2022refined}; (n) CT-crackseg \cite{tao2023convolutional}. The true-positive, false-positive, and false-negative pixels are shown in green, blue, and red, respectively.}
		\label{fig_com}
	\end{center}
\end{figure*}

\begin{figure*}[!t]
	\begin{center}
		\includegraphics[width=\textwidth]{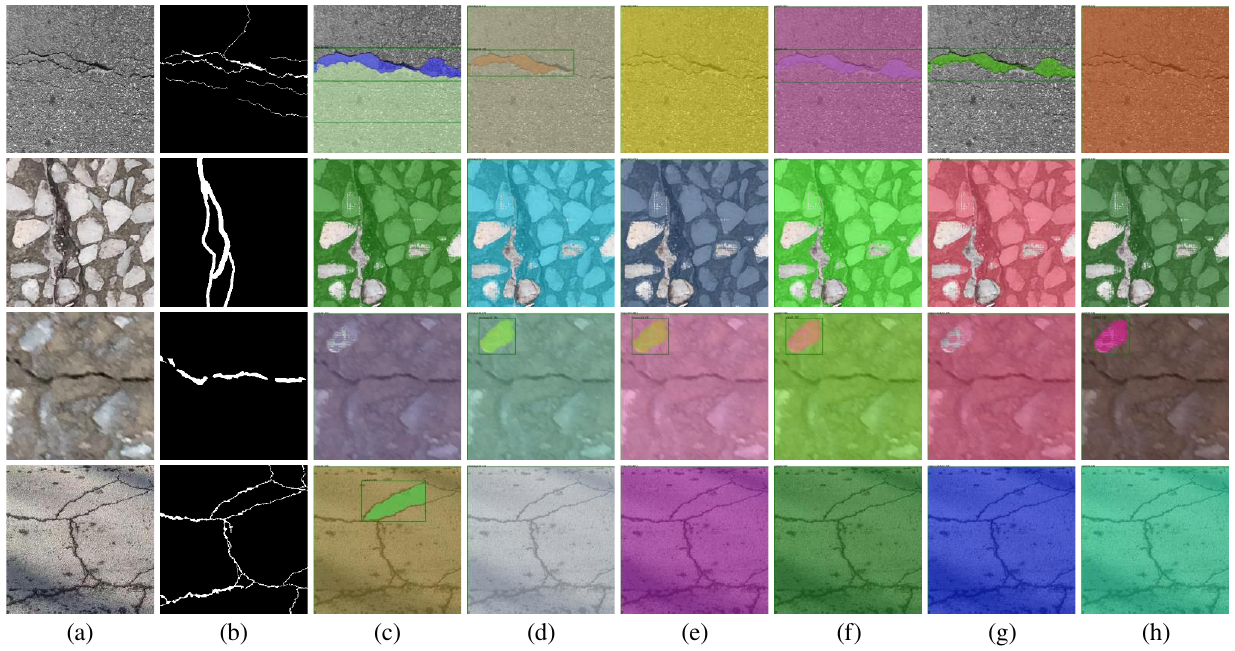}
		\centering
        \captionsetup{font={small}}
		\caption{Examples of Grounded-SAM \cite{ren2024grounded} on the AigleRN \cite{amhaz2016automatic}, CrackNJ156 \cite{xu2022pavement}, and proposed UDTIRI-Crack datasets: (a) Input road image; (b) Pixel-level label; (c) Results with text prompt ``crack"; (d) Results with text prompt ``damage"; (e) Results with text prompt ``fissure"; (f) Results with text prompt ``gap"; (g) Results with text prompt ``road crack"; (h) Results with text prompt ``split";
        }
		\label{fig_LLM}
	\end{center}
\end{figure*}

The public general-purpose supervised methods utilized for comparison include four CNN-based (BiseNet \cite{yu2018bisenet}, PSPNet \cite{zhao2017pyramid}, LEDNet \cite{wang2019lednet}, DeepLabv3+ \cite{chen2018encoder}) and six Transformer-based (LM-Net \cite{lu2024lm}, SwinTransformer \cite{liu2021swin}, SegFormer \cite{xie2021segformer}, TransUnet\cite{chen2021transunet}, SCTNet \cite{xu2024sctnet}, AFFormer \cite{dong2023head}) models. The public road 
crack detection-specific supervised methods utilized for comparison include seven CNN-based (ECSNet \cite{zhang2023ecsnet}, Deepcrack18 \cite{zou2018deepcrack}, Deepcrack19 \cite{liu2019deepcrack}, SCCDNet \cite{li2021sccdnet}, Crack-Att \cite{xu2023crack}, CDLN \cite{manjunatha2024crackdenselinknet}, SegDecNet++ \cite{tabernik2023automated}), and two Transformer-based (LECSFormer \cite{chen2022refined}, CT-crackseg \cite{tao2023convolutional}) models. The comparative results of these methods on the proposed UDTIRI-Crack dataset are presented both quantitatively and qualitatively in Table \ref{table_UDTIRI-Crack}, Fig. \ref{fig_UDTIRI-Crack}, and Fig. \ref{fig_com}, respectively. The results clearly demonstrate that crack detection-specific methods tend to outperform general-purpose methods, primarily due to the incorporation of novel network modules specifically tailored to address the unique challenges of road crack detection. Specifically, Crack-Att \cite{xu2023crack} and CT-crackseg \cite{tao2023convolutional} achieve the highest detection performance among crack detection-specific CNN-based and Transformer-based methods, respectively, with $5.646-10.037\%$ and $0.640-8.829\%$ improvement in IoU compared to general-purpose CNN-based and Transformer-based methods, respectively. The results reveal the effectiveness of the designed parallel attention mechanism and multi-scale feature map merging technique from Crack-Att, as well as the designed dilated residual blocks and boundary awareness module from CT-crackseg for refined road crack detection performance.
Notably, CT-crackseg achieves an improvement of $3.518\%$, $3.014\%$, $1.864\%$ in IoU, F1-Score, and AIoU over Crack-Att, further demonstrating the superiority of the hybrid CNN-Transformer architecture. This architecture effectively integrates detailed hierarchical spatial information with global long-range contextual information, enabling more precise and robust crack detection results. 

Apart from detection capability, the model's resource consumption and computational complexity are also crucial considerations when embedding it into intelligent road inspection vehicles. To quantitatively evaluate these aspects, we use model parameters and FPS as metrics. 
The results presented in Fig. \ref{fig_UDTIRI-Crack} indicate that ECSNet \cite{zhang2023ecsnet} has the smallest model size with $0.256$ million parameters, while Deepcrack19 \cite{liu2019deepcrack} achieves the highest FPS at $276.695$. Nonetheless, none of the evaluated public models successfully balance road crack detection accuracy, model size, and real-time performance.

To further assess the generalizability of the compared methods, we evaluated their detection performance on the AigleRN \cite{amhaz2016automatic} and CrackNJ156 \cite{xu2022pavement} datasets, with all models trained on the proposed UDTIRI-Crack dataset. The quantitative results, presented in Table \ref{table_aiglern} and Table \ref{table_cracknj}, indicate that methods specifically designed for crack detection generally exhibit better generalizability than general-purpose methods. This suggests that the specialized modules developed for this task are more effective at capturing the feature information of road cracks. Notably, the results in Table \ref{table_aiglern} demonstrated the robustness of LM-Net \cite{lu2024lm} in road crack detection under shadow conditions. However, on the CrackNJ156 dataset, the top-performing CT-crackseg model achieved an IoU of only $28.631\%$. This is attributed to the variations in weather, season, and lighting conditions between the images in CrackNJ156 and those in UDTIRI-Crack. These findings highlight that existing supervised road crack detection algorithms still lack robust generalizability across diverse scenarios.

\section{Existing challenges and future trends}
\label{sec.future}

\subsection{Insufficient Dataset Amount and Label Quality}

Although there exist some public datasets for road crack detection methods, compared with public natural scene datasets, such as more than 14 million sample data in the ImageNet dataset \cite{deng2009imagenet}, the amount and diversity of them is far from sufficient. As supervised deep learning-based road crack detection models demand a substantial quantity of labeled data to learn abundant feature information, establishing large-scale labeled road crack datasets is an existing challenge. Advanced image generation methods, such as generative adversarial networks (GAN) \cite{goodfellow2020generative} and diffusion models \cite{ho2020denoising}, can be introduced to produce manifold and excessive road crack data. In addition, the annotation quality of existing datasets varies greatly, especially for pixel-level annotations used in semantic segmentation-based methods, which affects the stability and sustainability of model training. Combining automatic annotation tools with manual refinement will be better to provide high-quality labels. Furthermore, establishing online benchmarks covering diverse scenarios, road surface materials, and crack types, is an imperative contribution to the development of road safety assessment research in the future. 

\subsection{Inadequate Detection and Real-Time Performance}

From the comparison results in Table \ref{table_UDTIRI-Crack} and Fig.\ref{fig_com}, it can be seen that even the best-performing public method still struggles to accurately detect road cracks, with evident instances of false positives and missed detections. Based on the experimental results and literature review, three key research directions for model improvement can be identified: (1) develop CNN-Transformer hybrid architectures that effectively integrate the detailed hierarchical spatial information captured by CNN with the global long-range contextual information provided by Transformer blocks; (2) design boundary refinement modules to improve the model's ability to accurately detect and delineate crack boundaries; (3) propose multi-scale feature extraction and prediction fusion training strategies to enhance the network's robustness and adaptability. Moreover, as illustrated in Fig. \ref{fig_UDTIRI-Crack}, there is a notable absence of public methods that achieve an optimal balance between road crack detection accuracy and real-time performance. A promising research direction lies in developing lightweight models that maintain relatively high detection performance while reducing model parameters and enhancing processing speed. Such advancements would reduce computational resource requirements and improve cost-effectiveness, thereby ensuring the economic feasibility of automatic road crack detection technology. 

\subsection{High-Dependence on Fine-annotated Dataset}

Most existing cutting-edge road crack detection methods remain predominantly supervised, and the training process relies on massive human-annotated labels. The creation of such fine labels demands professional expertise and entails significant labor and time expenditure. To fill this gap, it is crucial to develop label-efficient road crack detection methods that substantially reduce annotation efforts and costs. While existing relevant methods have shown some promise, there remains a need for more advanced and comprehensive solutions. For example, unsupervised methods such as \cite{ma2024up} face challenges in accurately detecting tiny cracks and can be disrupted by small anomalies. Weakly-supervised methods \cite{konig2022weakly}, \cite{al2023weakly} require manual parameter tuning, which limits their practical applicability. 

\subsection{Limited Model Generalizability}

\begin{figure}[!t]
	\begin{center}
	\centering
	\includegraphics[width=0.49\textwidth]{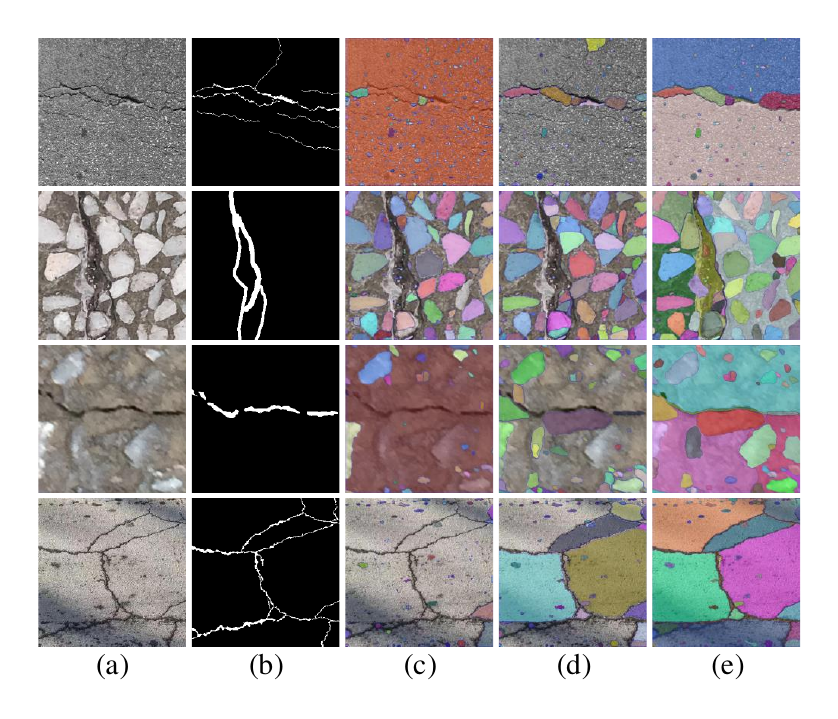}
	\centering
    \captionsetup{font={small}}
	\caption{Examples of SAM series results on the AigleRN \cite{amhaz2016automatic}, CrackNJ156 \cite{xu2022pavement}, and proposed UDTIRI-Crack datasets: (a) Input road image; (b) Pixel-level label; (c) Results from SAM \cite{kirillov2023segment}; (d) Results from SAM$2$ \cite{ravi2024sam}; (e) Results from semantic SAM \cite{li2023semantic}.}
		\label{fig_sam}
	\end{center}
\end{figure}

The comparison results in Table \ref{table_aiglern} and Table \ref{table_cracknj} indicate that, although crack detection-specific supervised methods demonstrate better generalizability compared to general-purpose supervised methods, their performance significantly deteriorates in the presence of scene variations and noise interference. This limitation is primarily attributed to the insufficient size of available datasets and the inherent dependence of supervised training methods on labeled data. To address this limitation, foundation models and LLMs (such as SAM \cite{kirillov2023segment} series and Grounded-SAM \cite{ren2024grounded}), which demonstrate strong generalizability across diverse image types, present a promising direction. We evaluate SAM, SAM$2$ \cite{ravi2024sam}, and semantic SAM \cite{li2023semantic} on the AigleRN \cite{amhaz2016automatic}, CrackNJ156 \cite{xu2022pavement}, and proposed UDTIRI-Crack datasets, respectively. Also, we evaluate the performance of Grounded-SAM on the three datasets by employing six distinct textual prompts: ``crack", ``damage", ``fissure", ``gap", ``road crack", and ``split". However, the qualitative results shown in Fig.\ref{fig_LLM} and Fig. \ref{fig_sam} indicate that directly introducing the existing foundation models and LLMs fails to achieve the desired road crack detection results, instead producing segmentation results limited to objects or road areas within the images, or even interpreting the entire input image as the output. This highlights the necessity of fine-tuning the SAM series and Grounded-SAM to adapt them specifically for road crack detection tasks. Furthermore, integrating label-efficient training strategies with sufficient road data generated by advanced image synthesis techniques can enhance the adapted models, enabling more robust and generalized detection performance for road cracks.

\section{Conclusion}
\label{sec.conclusion}

This paper provides a comprehensive review of SoTA deep learning-based road crack detection algorithms, from single-modal supervised methods to the emerging realm of data-fusion supervised, and label-efficient methods. We propose UDTIRI-Crack, the first extensive online benchmark in this field, alongside the AigleRN and CrackNJ156 datasets to evaluate the detection performance, generalizability, and computational efficiency of public SoTA semantic segmentation-based algorithms for road crack detection. 
The experimental results indicate that a promising research direction involves developing lightweight models that maintain relatively high detection performance while reducing model complexity and enhancing processing speed. Given the reliance of supervised semantic segmentation-based methods on massive human-annotated datasets, the creation of which demands significant labor and time costs, designing advanced label-efficient road crack detection methods becomes an escalating demand.
Moreover, leveraging foundation models and LLMs for their robust generalizability in this field necessitates further adaptation and the development of tailored training strategies.

\bibliographystyle{IEEEtran}

\bibliography{main}
\IEEEauthorblockN{\textbf{Nachuan Ma}} 
\IEEEauthorblockA{is currently pursuing his Ph.D. degree at Tongji University with a research focus on visual perception techniques for autonomous driving.}\\
\newline
\IEEEauthorblockN{\textbf{Zhengfei Song}} 
\IEEEauthorblockA{is currently pursuing his B.E. degree at Tongji University with a research focus on computer vision and deep learning.}\\
\newline
\IEEEauthorblockN{\textbf{Qiang Hu}} 
\IEEEauthorblockA{is currently pursuing his B.E. degree at Tongji University with a research focus on computer vision techniques for autonomous driving.}\\
\newline
\IEEEauthorblockN{\textbf{Chuang-Wei Liu}} 
\IEEEauthorblockA{is currently pursuing his Ph.D. degree at Tongji University with a research focus on computer vision techniques for autonomous driving.}\\
\newline
\IEEEauthorblockN{\textbf{Yu Han}} 
\IEEEauthorblockA{ is currently pursuing his M.S. degree at Donghua University with a research focus on computer vision.}\\
\newline
\IEEEauthorblockN{\textbf{Yanting Zhang}}
\IEEEauthorblockA{(Member, IEEE) received her Ph.D. degree in the School of Information and Communication Engineering from Beijing
University of Posts and Telecommunications in 2020. She is currently an Associate Professor in the School of Computer Science and Technology at Donghua University. Her research interests include computer vision and video/image processing.}\\
\newline
\IEEEauthorblockN{\textbf{Rui Fan}}
\IEEEauthorblockA{(Senior Member, IEEE) received his Ph.D. degree in Electrical and Electronic Engineering from the University of Bristol in 2018. He is currently a Full Professor with the College of Electronics \& Information Engineering and the Shanghai Research Institute for Intelligent Autonomous Systems at Tongji University. His research interests include computer vision, deep learning, and robotics.}\\
\newline
\IEEEauthorblockN{\textbf{Lihua Xie}}
\IEEEauthorblockA{(Fellow, IEEE) received the Ph.D. degree in Electrical Engineering from the University of Newcastle, Australia, in 1992. Since 1992, he has been a Professor and Director at the Center for Advanced Robotics Technology Innovation in the School of Electrical and Electronic Engineering at Nanyang Technological University, Singapore. Dr. Xie’s research interests include robust control and estimation, networked control systems, multi-agent networks, and unmanned systems.}

\end{document}